\documentclass[a4paper,11pt]{article}

\usepackage{fourier}
\usepackage{color}
 \usepackage{graphicx}
\usepackage{url}
\usepackage[affil-it]{authblk}
\usepackage{amsmath}
\usepackage{wrapfig}
\usepackage{xspace}

\usepackage[T1]{fontenc}
\usepackage{times}

\usepackage{times}
\usepackage{epsfig}
\usepackage{graphicx}
\usepackage{amsmath}
\usepackage{amssymb}
\usepackage{textcomp}
\usepackage{multirow}
\usepackage{adjustbox}

\usepackage{multirow}
\usepackage{colortbl}
\usepackage{hhline}
\usepackage{subcaption}
\usepackage{verbatim}
\usepackage{adjustbox}

\usepackage{gensymb}
\usepackage{flushend}
\usepackage{xcolor}
\usepackage{hyperref}

\usepackage{expl3}
\ExplSyntaxOn
\newcommand\latinabbrev[1]{
  \peek_meaning:NTF . {
    #1\@}%
  { \peek_catcode:NTF a {
      #1.\@ }%
    {#1.\@}}}
\ExplSyntaxOff

\def\etc{\latinabbrev{etc}}

\begin{document}

\title{FisheyeMultiNet: Real-time Multi-task Learning Architecture for Surround-view Automated Parking System}

\author{Pullarao Maddu, Wayne Doherty, Ganesh Sistu, Isabelle Leang, Michal Uricar, Sumanth Chennupati, Hazem Rashed, Jonathan Horgan, Ciaran Hughes and Senthil Yogamani}
\affil{Valeo Vision Systems, Ireland}
\date{}
\maketitle
\thispagestyle{empty}

\vspace{-1.2cm}

\begin{abstract}
Automated Parking is a low speed manoeuvring scenario which is quite unstructured and complex, requiring full 360{\degree} near-field sensing around the vehicle. In this paper, we discuss the design and implementation of an automated parking system from the perspective of camera based deep learning algorithms. We provide a holistic overview of an industrial system covering the embedded system, use cases and the deep learning architecture. We demonstrate a real-time multi-task deep learning network called FisheyeMultiNet, which detects all the necessary objects for parking on a low-power embedded system. FisheyeMultiNet runs at 15 fps for 4 cameras and it has three tasks namely object detection, semantic segmentation and soiling detection. To encourage further research, we release a partial dataset of 5,000 images containing semantic segmentation and bounding box detection ground truth via WoodScape project \cite{yogamani2019woodscape}.
\end{abstract}


\section{Introduction}
\begin{wrapfigure}{r}{0.55\textwidth}
  \vspace{-20pt}
  \begin{center}
    \includegraphics[width=0.5\columnwidth]{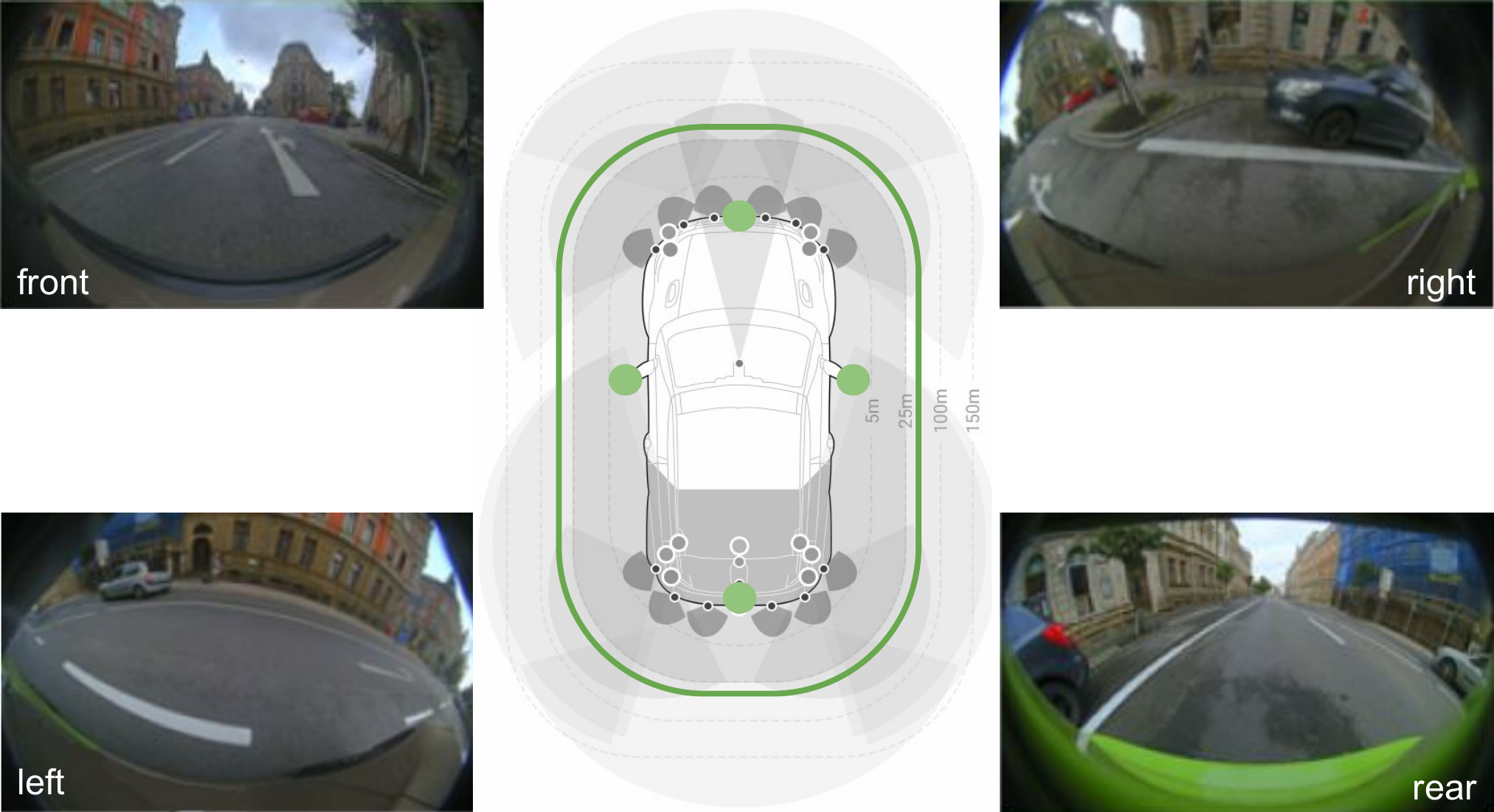}
\end{center}
\vspace{-20pt}
  \caption{Images from the surround-view camera network showing near field sensing and wide field of view.}
\label{fig:svs}
  \vspace{-10pt}
\end{wrapfigure}

Recently, Autonomous Driving (AD) gained huge attention with significant progress in deep learning and computer vision algorithms \cite{Klette2017}, where it is considered one of the highly trending technologies all over the globe. Within the next 5-10 years, AD is expected to be deployed commercially. Currently, most of the automotive original equipment manufacturers (OEMs) over the world are working on development projects focusing on AD technology \cite{ro2019factor}. The complexity of the system must be acceptable for the purpose of producing commercial cars which adds limitations to the hardware used for production. 
Fisheye cameras offer a distinct advantage for automotive applications. Given their extremely wide field of view, they can observe the full surrounding of a vehicle with a minimal number of sensors. Typically four cameras is all that is required for full 360$^\circ$ coverage of a car (Figure \ref{fig:svs}). Nevertheless, this advantage comes with some drawbacks in the significantly more complex projection geometry that fisheye cameras exhibit.
This advantage comes with a cost in the significantly more complex projection geometry exhibited by fisheye cameras. 

Convolutional neural networks (CNNs) have became the standard building block for the majority of visual perception tasks in autonomous vehicles. Bounding boxes for object detection is one of the first successful applications of CNNs for detecting not only pedestrians and vehicles, but also their positions. Recently semantic segmentation is becoming more mature \cite{siam2017deep} \cite{siam2018rtseg}, starting with detection of roadway objects like road surface, lanes, road markings, curbs, etc. CNNs are also becoming competitive for geometric vision tasks like depth estimation \cite{kumar2018near}, Visual SLAM \cite{milz2018visual}, etc. Despite rapid progress in the computational power of embedded systems and of specialized CNN hardware accelerators, real-time performance of semantic segmentation 
is still challenging. In this paper, we focus on deep learning architecture for an automated parking system which is relatively less explored in the literature \cite{heimberger2017computer}.

The rest of the paper is structured as follows. Section \ref{sec:parking} provides an overview of parking system use cases and necessary visual perception modules.  Section \ref{sec:sysarch} details a concrete implementation of efficient multi-task architecture with results and discusses how it fits into the overall system architecture. 
Finally, Section \ref{sec:conc} summarizes the paper and provides potential future directions.

\section{Automated Parking System} \label{sec:parking}


\subsection{Parking Use cases}

\begin{figure}[!ht]
\centering
\includegraphics[width=.85\columnwidth]{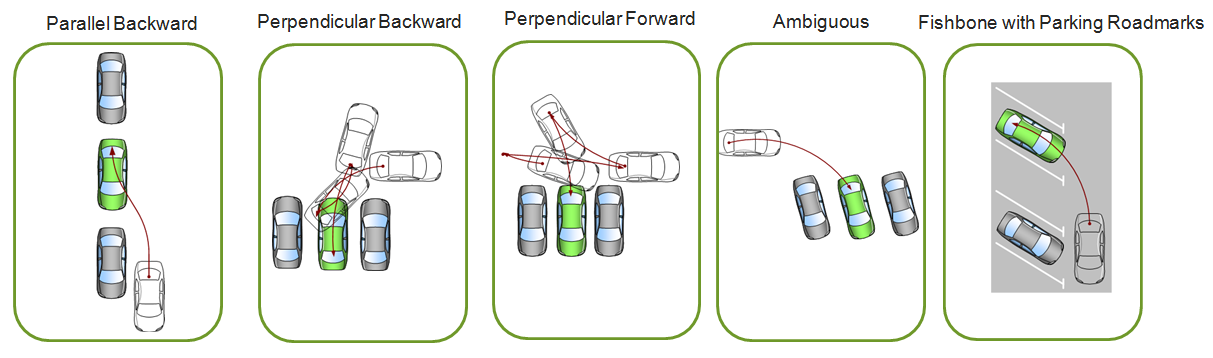}
\caption{Classification of Parking scenarios - (a) Parallel Backward Parking (b) Perpendicular Backward Parking (c) Perpendicular Forward Parking (d) Ambiguous Parking and (e) Fishbone Parking with roadmarkings. }
\label{fig:parking-classification}
\end{figure}

\indent \textbf{Parallel parking:} The system attempts to align the vehicle in parallel to the curb or the road as illustrated in \ref{fig:parking-classification}(a). In such a strategy, the vehicle usually parks in one maneuver, and further maneuvers are required for alignment with curb and the vehicles around. Robust object detection and curb classification has to be implemented to minimize the distance between the vehicle and the curb and ensure the vehicles in front and behind are avoided. Conventional ultrasonic sensors are capable of detecting curbs, however fusion with cameras greatly enhances the classification and position accuracy. 

\textbf{Perpendicular parking:} The system tries to find a lateral parking slot, where the width of the slot is sufficient for the vehicle, with additional room for opening the doors and safety distances. If the slot is found to fit the required size, then a trajectory that minimizes the number of maneuvers necessary is planned to reach the slot target. This parking strategy can be performed in backward direction as illustrated in Figure~\ref{fig:parking-classification}(b) or forward direction as shown in Figure~\ref{fig:parking-classification}(c). Ultrasonic sensors are quite unreliable in the detection of other vehicle's corners due to missing and incorrect reflections of the ultrasonic waves,resulting in the multiple re-measurements to improve the detection. This may result in some additional maneuvers to overcome the error introduced from using ultrasonic sensors only. As well as this ultrasonics are only useful in parking between two objects, being unable to detect road markings. Fusion with a camera sensor provides improved performance in multiple aspects. For instance, computer vision techniques can provide complementary information for depth estimation using Structure from Motion (SFM). Cameras are also able to detect the white line markings which allow for detection of slots where there are multiple empty slots in a group.

\textbf{Ambiguous Parking:} This parking scenario is neither parallel or perpendicular. The orientation must be detected from the surrounding vehicles as in Figure~\ref{fig:parking-classification}(d). Due to the increased detection range, and the complete sensor coverage around the vehicles that cameras provide, computer vision provides a more appropriate reaction of the ego-vehicle in such situations. For instance, ultrasonic sensors do not provide information about the ego-vehicle's flank, objects have to be tracked blindly in that area using the vehicles motion, while this information is provided in a 360 surround-view while using fisheye cameras. By using the complementary color information provided by cameras, systems will also be able to detect any suddenly occurring objects with higher confidence and thus react in a more timely manner compared to ultrasonics alone. 

\textbf{Fishbone Parking:} Figure \ref{fig:parking-classification}(e) shows an example of fishbone parking where there is a huge limitation in ultrasonic sensors. To be able to detect the slot orientation using ultrasonic sensors only, the vehicle has to drive inside the slot to detect the orientation from the surrounding vehicles, as the density of reflections is too low when the vehicle is outside the slot. Therefore, detection of such a slot during the search phase is not possible. Fusion with camera enables an increased range of detection using both object detection and slot marking detection. This use case cannot be covered using ultrasonic sensors solely. 

\textbf{Home Parking:} Thanks to the huge progress in computer vision and self-parking technology, higher-level applications have been introduced for more comfort and better driving experiences. One of which is "Home Parking" where the system is trained by the driver to follow a set trajectory and park in a particular spot. The surrounding area is stored on the system and particular landmarks recorded. By doing this the vehicle is capable of localizing itself within the environment in future and driving completely autonomously onto the stored trajectory and following it to it's regular parking space. 

\textbf{Valet Parking:} Significant progress has been made in automated parking even without a stored trajectory. In this case, the system is completely autonomous in it's slot-search, selection, and parking without having any prior knowledge about the environment or a predefined trajectory. 


\subsection{Necessary Vision Modules}\label{subsec:necessary_vision_modules}

\textbf{Parking slot detection:} The first and foremost step in automated parking is the selection of a valid parking space, in which a car can be safely parked. An ideal parking slot detection algorithm shall detect several types of parking slots, as shown in Figure~\ref{fig:parking-classification}. Parking slot detection can be further broken down into several stages. It involves detection of line markings, curbs, vehicles, shrubs and walls as all of these are necessary in recognizing an open parking slot. Additionally, it is of vital importance an accurate measurement of the width and length of the slot can be made to ensure the vehicle can safely fit within. 

\textbf{Freespace detection:} The final objective of autonomous parking system or complete autonomous driving systems is navigating the car to a target. Therefor the freespace (area free of pedestrians, vehicles, cyclists or any other objects that have potential risk of damage or injury while passing over them) or "driveable" area information is critical. Such information is also crucial in situations when evasive maneuvers are needed in real time to minimize the risk of collision. 

\textbf{Pedestrian detection:}
Collision risk usually arises from object classes that can be moving. One of such classes is the pedestrian class. Pedestrian detection comprises a challenging task due to several reasons. For instance, they are very difficult to track because pedestrian motion can be erratic and difficult to predict. A pedestrian may suddenly appear behind a vehicle while attempting to park. Knowing the object belongs to the pedestrian class, the system should expect it to move away, and thus should not abort at that moment. Pedestrian classification is very helpful in other autonomous driving situations as well, e.g. a child suddenly crosses the street and the vehicle has to suddenly brake. Infrared cameras can be utilized to maximize the performance of pedestrian detection systems, due to their capability to capture thermal energy \cite{Baek2017sensors}, but this can be costly in production systems. 

\textbf{Vehicle detection:} 
Vehicle detection is one of the most important automotive computer vision tasks. It is very helpful in the scope of autonomous parking for many reasons.  
For example, the ability to distinguish between high obstacles, such as shrubs or walls and vehicles. In a parking situation it is of vital importance the system can recognize a vehicle which has the ability to move and obstruct the planned trajectory of our car, and a wall which we plan to park alongside, knowing it will be stationary throughout our manoeuvre.  
Typically, in the AD scenario, the system has to react to dynamic vehicles surrounding the ego-vehicle. Such vehicles have to be tracked to avoid suddenly occurring vehicles after occlusion. The first step to perform such a task is vehicle classification. 

\textbf{Cyclist detection:} 
Cyclists can be classified as pedestrians. However, cyclists have the ability to move faster with less maneuverability. Thus, distinguishing between cyclists and pedestrians provides additional information for the system that helps in tracking such objects. 

\textbf{Soiling Detection:} 
Cameras embedded within the vehicles are directly exposed to an external environment and there is a good chance that they get soiled due to bad weather conditions such as rain, fog, snow, etc \cite{uricar2019soilingnet}. Moreover, dust and mud have a strong affect of degraded computer vision performance. Compared to other types of sensors, cameras have much higher degradation in performance due to soiling. Thus, it is critical to robustly detect soiling on the cameras, especially for higher levels of autonomous driving. Soiling detection was first implemented to alarm the driver that there will be degraded performance in the environment perception system. In a high-level autonomous system there could be fatal consequences if information from soiled cameras is relied on, without having prior information that it is not correct.

\section{Parking System Architecture} \label{sec:sysarch}

\subsection{Overall Software Architecture}

\begin{figure}[!ht]
  \begin{center}
\includegraphics[width=0.8\textwidth]{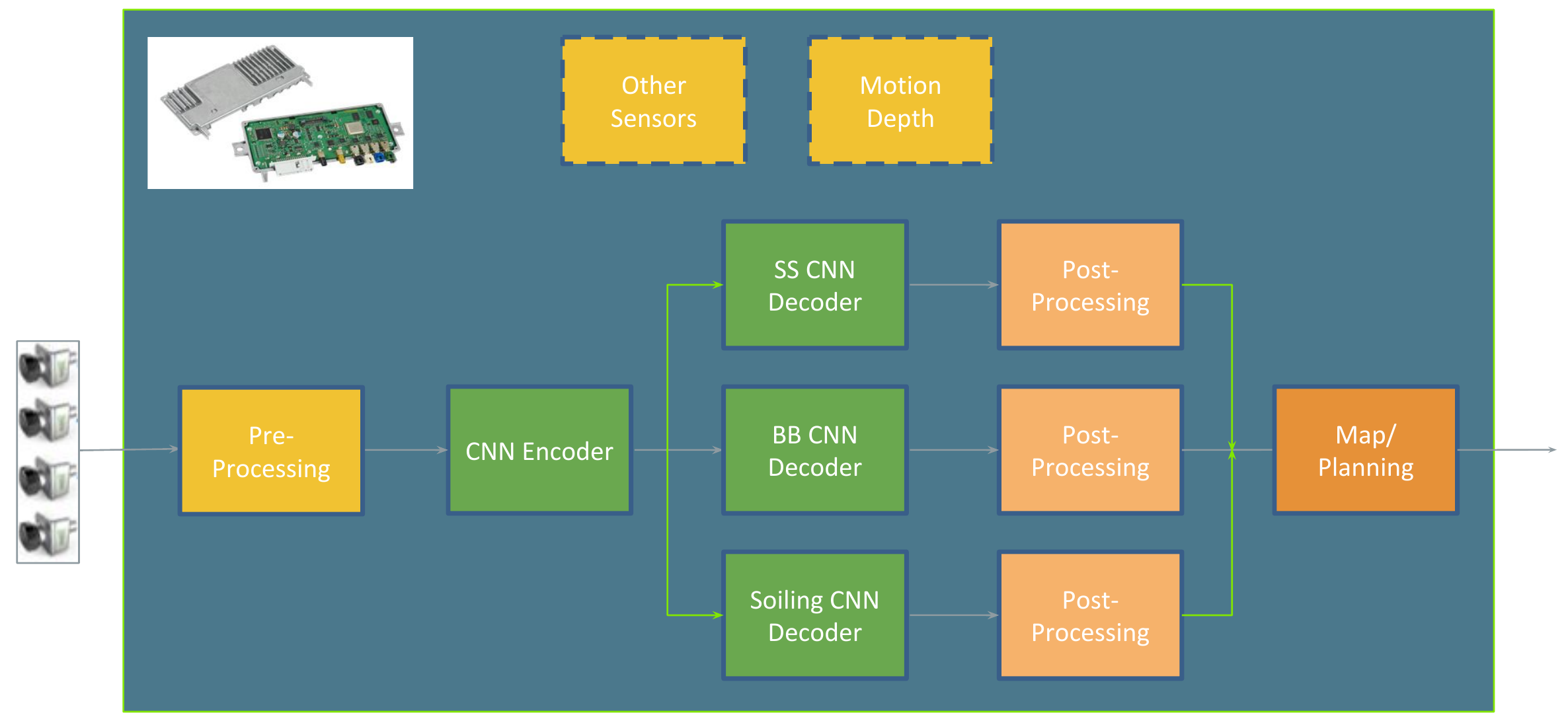}
\end{center}
\vspace{-20pt}
\caption{Parking System Architecture}
\label{fig:sysarch}
\vspace{-10pt}
\end{figure}

The block diagram of our system is illustrated in Figure \ref{fig:sysarch}. The first step in an industrial system is the SOC (System on Chip) selection for embedded systems, based on criteria including performance (Tera Operations Per Second (TOPS), utilisation, bandwidth), cost, power consumption, heat dissipation, high to low end scalability and programmability. The SOC choice provides the computational bounds in the design of algorithms. A typical embedded system is shown on top left of the block diagram. In computer vision, deep learning is playing a dominant role in various recognition tasks and gradually for geometric tasks, like depth and motion estimation also. The progress in CNN has also led to the hardware manufacturers including a custom hardware intellectual property core to provide a high throughput of over $10$ TOPS. The current system we are developing our algorithms on, has $1$ TOPS of compute power, consuming less than $10$ watts of power. 

The necessary object detection modules were discussed in Section~\ref{subsec:necessary_vision_modules}. In previous systems, some modules, for instance pedestrian detection, was done using machine learning techniques while others, like parking slot detection were done using classical computer vision techniques. Due to recent advancements in deep learning, all of the necessary vision modules can now be done using deep learning models. Thus, we propose a unified multi-task architecture for doing all these tasks, that runs on a Hardware accelerator (Green in the block diagram (Fig.~\ref{fig:sysarch})). This will be discussed in more detail in the next section. The deep learning model provides necessary functionality for parking. However, to add robustness, additional cues like motion estimation and depth estimation can be used along with other sensors like Ultrasonics, Radar, \etc. In this paper, we focus on the basic solution for a parking system using deep learning only. Any detected objects from the four cameras are recorded in image coordinates, mapped to world coordinates to create a common representation and fed into a virtual map to plan maneuvering of the car for automated parking. Road markings and curbs are handled in the same way, also being sent to the map building a viable model for the world around us. Bounding boxes can be established around objects such as pedestrians and vehicles by assuming a flat ground plane and mapping the foot-point (intersection of object to ground plane) to a world position using the vehicle and camera calibration. Depth estimation can handle cases where the foot-point is occluded or the road is no flat. 

\subsection{Proposed Multi-task Architecture}

Various visual perception tasks like semantic segmentation~\cite{paszke2016enet}, bounding object detection \cite{redmon2016you}, motion segmentation~\cite{siam2018modnet}, depth estimation and soiling detection are commonly addressed using 
an encoder-decoder style architecture in deep learning. Many works have focused on solving these tasks independently. However, multi-task learning \cite{sistu2019neurall,visapp19,8500504} enables the solving of these tasks using a single model. The main advantage of a multi-task network is its high computational efficiency, which is most suitable for a low cost embedded device. In a simple scenario, where a multi-task network solving two tasks using a common encoder that shares $30 \%$ of common load is comparatively much better than independent networks consuming the whole processing power available without common load sharing. In this case, an additional task can also be solved with remaining computing resources. This, in fact, offers scalability for adding new tasks at a minimal computation complexity. \cite{chennupati2019} provided a detailed overview on negligible incremental computational complexity while increasing number of joint tasks solved by a multi-task network. On the other hand, using pre-trained encoders (say  ResNet~\cite{he2016deep}) as a common encoder stage in multi-task networks reduces training time and alleviates the daunting requirements of massive data to optimize. Reusing the encoder also provides regularization across different tasks. \\

\begin{figure}[!ht]
    \centering
    \vspace{-0.2cm}
    \includegraphics[width=.7\textwidth]{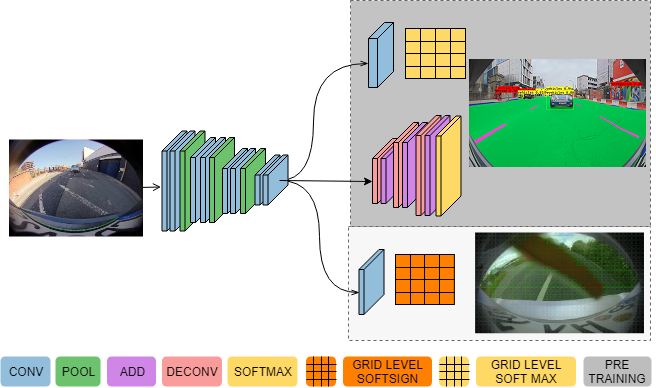}
    \caption{Illustration of FisheyeMultiNet architecture comprising of object detection, semantic segmentation and soiling detection tasks.}
    \label{fig:multi-stream-task}
\end{figure}

\textbf{Network Architecture:} We propose a multi-task network called FisheyeMultiNet, having a shared encoder and three independent decoders that perform joint semantic segmentation, object detection and soiling detection as shown in Figure~\ref{fig:multi-stream-task}. A semantic segmentation decoder provides valuable lane markings, road and sidewalk information, while an object detection decoder provides bounding boxes of pedestrians, cyclists, vehicles, etc. These two tasks primarily provide solutions to the major vision modules discussed in Section~\ref{sec:parking}. A soiling detection decoder outputs the presence of external contamination on the camera lens, providing classification per tile for obtaining the localization of soiling in the image. 
We treat the camera soiling detection task as a mixed multilabel-categorical classification problem focusing on a classifier, which jointly classifies a single image with a binary indicator array, where each $0$ or $1$ corresponds to a missing or present class respectively, and simultaneously assigns a categorical label. The classes to detect are $\{ \mathrm{opaque}, \mathrm{transparent} \}$. Typically, opaque soiling arises from mud and dust, and transparent soiling arises from water and ice.

The raw fisheye images are passed to a common encoder built using the ResNet10~\cite{he2016deep} encoder. This encoder is pre-trained on ImageNet~\cite{ILSVRC15} and then trained on raw fisheye WoodScape images. The semantic segmentation network is built using the FCN8~\cite{long2015fully} decoder with skip connections from the ResNet10 encoder. The object detection decoder is built using a grid level softmax layer, while the soiling decoder is built using a grid level softsign layer. The categorical cross entropy is used as a loss metric for semantic segmentation and soiling detection, while average precision is used as the loss metric to express individual task losses. The total loss of the network is expressed as a weighted arithmetic combination of individual task losses and optimized using the Adam~\cite{Kingma-B14} optimizer. We do this intending to have a drastic increase in memory available and computational efficiency with just a small reduction in accuracy.
We make use of several standard optimization techniques to further improve the runtime, and achieve $10$ fps for four cameras on an automotive grade low power SOC. Some examples are: (1) Reducing number of channels in each layer, (2) Reducing number of skip connections for memory efficiency, and (3) Restricting segmentation decoder to image below the horizon line (only for roadway objects). \\


\textbf{Datasets:} The development of our architecture was primarily done on our internal parking dataset, which originates from three distinct geographical locations: USA, Europe, and China. While the majority of data was obtained from saloon vehicles, there is a significant subset that comes from a sports utility vehicle (SUV) ensuring a strong mix in sensor mechanical configurations. It consists of four $1$ Megapixel RGB fisheye cameras ($190^\circ$ hFOV). After the collection of images, an instance selection algorithm is applied to remove redundancy \cite{michalvisapp19} and produce the final dataset which consists of $5,000$ samples. To the best of the authors' knowledge, this is the first public dataset for automated parking. The dataset is split into three chunks in a ratio of $6:1:3$, namely training, validation, and testing. This dataset and the baseline multi-task model will be made public to the research community via our WoodScape project \cite{yogamani2019woodscape}.

\subsection{Results and Discussion} \label{sec:results}

In this section, we explain the experimental settings including the datasets used, training algorithm details, etc. and discuss the results. We used our fisheye dataset comprising of $10,000$ images. 
We implemented our baseline object detection, semantic segmentation networks and our proposed multi-task network using Keras. 
All input images were resized to $1280 \times 384$ because of memory requirements needed for multiple tasks. Table~\ref{tab:mtl} summarizes the obtained results for the single task (STL) independent networks and multi-task (MTL) networks on our parking fisheye datasets. 

\begin{table*}[]
\small
\centering
\caption{Comparison Study: Single task vs. Multi-task FisheyeMultiNet}
\resizebox{0.45\columnwidth}{!}{%
\begin{adjustbox}{angle=0}
\begin{tabular}{lllllll}
\hline
Databases                                & Metrics           & STL Seg    & STL Det           & MTL     \\
\hline
\multirow{4}{*}{Parking Seg}     
                                         & JI road           & 0.9574          &                  & 0.9514          \\
                                         & JI lane           & 0.6517          &                  & 0.6424           \\
                                         & JI curb           & 0.5960           &                  & 0.5850          \\
                                         & \textbf{mean IOU} & \textbf{0.7350} & \textbf{}        & \textbf{0.7263} \\
\hline
\multirow{4}{*}{Parking Det}         & AP Vehicle            &                 & 0.6910                    & 0.7016         \\
                                         & AP person         &                 & 0.3620           & 0.3609          \\
                                         & AP cyclist         &                 & 0.3682           & 0.3817          \\                                         
                                         & \textbf{mean AP}  & \textbf{}       & \textbf{0.4737}  & \textbf{0.4814} \\
\hline
\multirow{2}{*}{Parking Soiling}         & TPR            &                 & 0.5581                    & 0.5532         \\
                                         & FPR         &                 & 0.1432           & 0.1443           \\
\hline
\end{tabular}
\end{adjustbox}
}
\label{tab:mtl}
\end{table*}

One of the main challenges of MTL is to balance the loss functions of all three tasks as the magnitude of the losses vary at different scales. This led to a faster convergence of certain tasks and divergence of other tasks. To handle this, we make use of a weighted loss function to normalize the losses. We update the task weights every epoch, based on loss gradients. We weigh the different tasks based on gradients observed after every epoch in a similar fashion to GradNorm~\cite{chen2017gradnorm}.  
We tested 3 configurations of the MTL loss, the first one (MTL) uses a simple sum of the segmentation loss and detection loss ($w_{seg}=w_{det}=1$). The two other configurations MTL$_{10}$ and MTL$_{100}$, use a weighted sum of the task losses where the segmentation loss is weighted with a weight $w_{seg}=10$ and $w_{seg}=100$ respectively. This compensates the difference of task loss scaling and $w_{seg}=100$ consistently improves the performance of the segmentation task for all the three datasets.
Experimental results show that performance of MTL networks are marginally lower than the STL networks. However, the computational gains offered by multi-task networks and a potential to improve performance by further fine-tuning, would make multi-task networks a more suitable option for future embedded deployment.

\section{Conclusion} \label{sec:conc}

In this paper, we provided a high level overview of a commercial grade automated parking system. We covered various aspects of the system in detail, including the embedded system architecture, parking use cases which need to be handled and the vision algorithms which solve these use cases. We have focused on a minimal system which can be designed via an efficient multi-task learning architecture using four fisheye cameras which provides $360{\degree}$ view surrounding the vehicle. We provided detailed quantitative results of the proposed deep learning architecture and show that the accuracy of an MTL network is not that much lower than an STL, despite the reduction in memory consumption and computational power. In addition, we released a dataset comprising of $5,000$ images with semantic segmentation \& bounding box annotation to encourage further research.

\vspace{-0.2cm}


\bibliographystyle{apalike}

\bibliography{imvip2019}

\end{document}